\documentclass[10pt,twocolumn,letterpaper]{article}

\usepackage{iccv}
\usepackage{times}
\usepackage{epsfig}
\usepackage{graphicx}
\usepackage{amsmath}
\usepackage{amssymb}

\usepackage{microtype}
\usepackage{subfigure}
\usepackage{booktabs} 
\usepackage{multirow}
\usepackage{flushend}

\newif\ifcomments
\commentsfalse
\commentstrue  
\ifcomments
\newcommand{\asma}[1]{\textcolor{red}{\bf\small [#1 --AG]}}
\else
\newcommand{\asma}[1]{}
\fi

\usepackage[breaklinks=true,bookmarks=false]{hyperref}

\iccvfinalcopy 

\def\iccvPaperID{****} 

\begin{document}


\title{Characterizing Sources of Uncertainty to Proxy Calibration and \\ Disambiguate Annotator and Data Bias}

\author{Asma Ghandeharioun\\
MIT\\
Cambridge, MA, US\\
{\tt\small asma\_gh@mit.edu}
\and
Brian Eoff, Brendan Jou\\
Google Research\\
New York City, NY, US\\
{\tt\small \{beoff, bjou\}@google.com}
\and
Rosalind W.~Picard\\
MIT\\
Cambridge, MA, US\\
{\tt\small picard@media.mit.edu}
}

\maketitle

\begin{abstract}

Supporting model interpretability for complex phenomena where annotators can legitimately disagree, such as emotion recognition, is a challenging machine learning task. In this work, we show that explicitly quantifying the \emph{uncertainty} in such settings has interpretability benefits. We use a simple modification of a classical network inference using Monte Carlo dropout to give measures of epistemic and aleatoric uncertainty. We identify a significant correlation between aleatoric uncertainty and human annotator disagreement ($r\approx.3$). Additionally, we demonstrate how difficult and subjective training samples can be identified using aleatoric uncertainty and how epistemic uncertainty can reveal data bias that could result in unfair predictions. We identify the total uncertainty as a suitable surrogate for model calibration, i.e.~the degree we can trust model's predicted confidence. In addition to explainability benefits, we observe modest performance boosts from incorporating model uncertainty.

\end{abstract}

\vspace{-.2cm}
\section{Introduction}

Supporting interpretability of an automated prediction system in complex tasks where human experts disagree is a challenging machine learning problem. In such settings, answering the following questions can help understand model's predictions: is the model uncertain due to capturing annotator biases and their subjective perspective? Or is it error-prone for a specific set of samples due to a distribution shift from the training data? Can the predicted confidence scores of the model be trusted? Do they represent the true likelihood so that we can intuit and reason about their results?

Emotion understanding is an icon for a learning setting where label ambiguity abounds. Most researchers agree that \textit{emotion} in itself is nuanced and the same input could be assigned different labels due to change in contextual information or the perspective of the reviewer \cite{frings2008trial}. Thus, disambiguating annotator and data bias and quantifying how well predictive confidence can be trusted is crucial to supporting explainability in emotion classification.

In this work, we extend beyond deterministic modeling of affect using Monte Carlo (MC) dropout \cite{gal2016dropout}, a technique that requires no changes to the neural network architecture and only minimal changes at inference time. This approach augments classification's per-class confidence scores with measures of uncertainty. We tease apart elements in the uncertainty estimates and investigate how each helps interpreting model predictions and its failure modes. We show this decomposition results in a proxy for inter-rater disagreement capturing annotators' bias, and a proxy highlighting bias in data that could potentially result in unfair predictions.

Given that humans have intuitive understanding of probability \cite{cosmides1996humans}, to support this intuition and provide interpretable confidence scores, calibration is required. That is, we expect predicted probability estimates to represent the true likelihood of correctness \cite{guo2017calibration}. Calibration can be seen as the degree of trust in predicted confidence scores of a classifier. We further investigate the relationship between uncertainty estimates and the degree of calibration, pinpointing samples where confidence predictions are (not) to be trusted.

While this technique helps interpretability by disambiguating sources of bias and relation to calibration, it can also boost performance. We report significant improvement in Jensen-Shannon divergence (JSD) between predicted and true class probabilities. We show a strong correlation between total uncertainty and JSD ($r\approx.6$), identifying it as a proxy for performance. We study the influence on accuracy, especially if given the option to reject classifying samples where the model lacks confidence.


To summarize, we use MC dropout with traditional neural network architectures and explore the benefits of resulting measures of uncertainty while disambiguating their source. Our contributions include: 1) introducing a proxy for inter-annotator disagreement, 2) demonstrating the power of such metrics in identifying difficult samples and bias in training data along with ways to alleviate them, 3) finding a surrogate for calibration, 4) showing improvements in performance in addition to interpretability benefits.

\section{Background \& Related Work}



Understanding what a model does not know is especially important to explain and understand its predictions. 
State-of-the-art classification results are mostly achieved by deep neural networks (DNN)---such as AlexNet, VGGNet, ResNet, etc.---that are deterministic in nature and not designed to model uncertainty.
Bayesian Neural Networks (BNN) have been an alternative to DNNs, providing a distribution over model parameters at an extra computational cost while increasing difficulty of conducting inference \cite{denker1991transforming, mackay1992practical}. These computational challenges hinder scalability of BNNs.

MC dropout \cite{gal2016dropout} has been introduced as an approximation of BNNs that can be achieved by keeping the same architecture of a deterministic DNN and only making minimal changes at inference time. Dropout, i.e.~randomly dropping weights at training time, is commonly used in DNNs as a regularization method. Drawing random dropout masks at test time can approximate a BNN. Recently, \cite{kendall2017uncertainties} demonstrated ways to additionally learn the observation noise parameter $\sigma$, thus modeling epistemic and aleatoric uncertainty in parallel. Epistemic uncertainty represents lack of confidence in one's knowledge attributed to missing information about the learning task. Aleatoric uncertainty is attributed to the stochastic behavior of observations. They evaluated the approach for use in regression tasks in depth estimation. A partial version of the model, only modeling aleatoric uncertainty, was evaluated for classification for semantic segmentation.


These efforts show great potential at empowering deterministic DNNs with Bayesian properties with negligible computational overhead. However, we will show complex, difficult tasks where reviewers disagree and data may not fully represent everyone, such as affect detection, can benefit from inferring different sources of uncertainty. Despite its importance, latent uncertainty quantification in emotion detection tasks is under-explored. However, there have been a few efforts regarding more realistic emotion recognition by incorporating explicit inter-annotator disagreement. For example, modeling \textit{perception} of uncertainty as measured by the standard deviation of labels captured from crowd-sourced annotations has been studied in  \cite{han2017hard}. While such efforts are valuable in affective computing applications, these approaches are supervised, are prone to error when annotations are sparse and varied in number, are not capable of capturing uncertainty introduced by model parameters or sources of noise other than human judgment. 


\section{Technical Approach}



The underlying architecture of our model is an Inception-ResNet-v1 \cite{szegedy2017inception} for extracting facial features, followed by a multi-layer perceptron for emotion classification. We built upon an open-source implementation \cite{facenetgithub} of FaceNet \cite{schroff2015facenet}.
We pre-trained the model up to the \texttt{Mixed-8b} layer using cross-entropy loss on face identity classes using the CASIA-WebFace dataset \cite{yi2014learning}. The pre-processing step included using a Multitask CNN \cite{zhang2016joint} to detect facial landmarks and extract facial bounding boxes in the form of 182$\times$182 pixel images. Since the utility of this training mechanism is to identify faces, it learns to ignore features that are invariant to one's identity, e.g.~facial expressions, in the later layers of the network while the earlier layers represent lower-level features. \texttt{Mixed-7a} best encoded and retained emotionally-relevant information based on our experiments (See \S\ref{sec:appendix-model}).

\subsection{Baseline}
After extracting features from layer \texttt{Mixed-7a}, a fully-connected network with two hidden layers was used to infer basic emotions. We refer to this model as \textit{Baseline}.
Facial Expression Recognition (FER) is an established emotion detection dataset \cite{goodfellow2013challenges}. FER+ is the same set of images, expanded to include at least 10 annotations from crowd-sourced taggers \cite{barsoum2016training}. We used FER+ train, private test, and public test subsets for training, hyper parameter tuning, and evaluation of our model performance, respectively. See \S\ref{sec:appendix-model} for details.

\begin{figure*}[!ht]
\centering
\includegraphics[width=0.85\columnwidth]{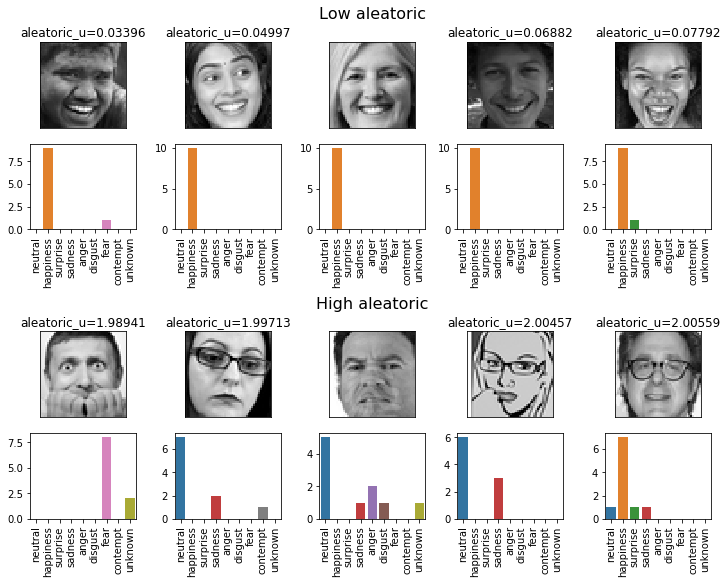}
\hspace{1.5cm}
\includegraphics[width=0.85\columnwidth]{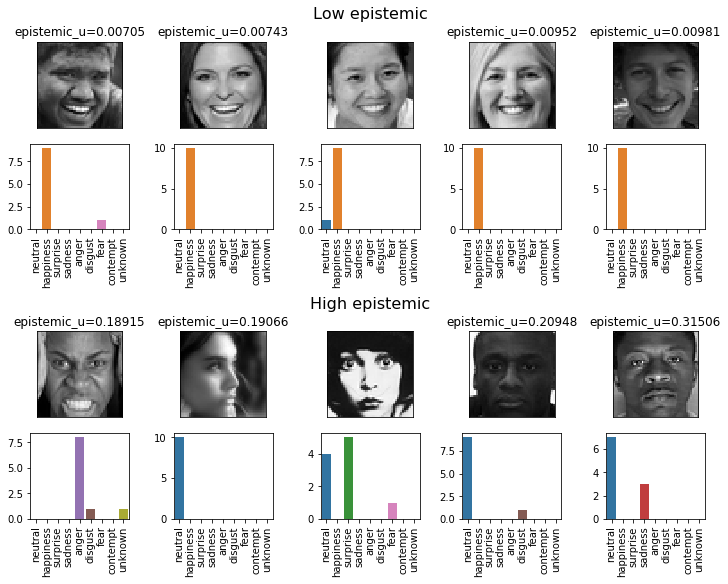}
  \caption{\textbf{Left}: Aleatoric uncertainty ($U_a$) - Samples with lowest $U_a$ are stereotypical expressions of emotion where annotators (almost) unanimously agree on the assigned label. Conversely, images with the highest $U_a$ either represent subjectivity involved in human annotations or low image quality, e.g. when the face is occluded by hands or the image is a drawing as opposed to a photograph. \textbf{Right}: Epistemic uncertainty ($U_e)$ - Samples with lowest $U_e$ show stereotypical expressions of emotion that are common in the training set. On the other hand, images with the highest $U_e$ include dark-skinned subgroups, a non-frontalized photo, and a highly illuminated image, even when there is near-perfect agreement across human-annotators. We believe this is due to the skewed pre-training dataset, suggesting that it is not equipped to encode such samples.}
  \label{fig:extreme-samples}
  \vspace{-.3cm}
\end{figure*}

\subsection{Epistemic \& Aleatoric Uncertainties}
For each input image, \textit{Baseline} predicts a length-$C$ logits vector $z$ which is then passed through a softmax operation to form a probability distribution $p$ over a set of class labels. For our new model, we move away from pointwise predictions, and put a Gaussian prior distribution over the network weights, $W\sim N(0,I)$. To overcome the intractability of computing the posterior distribution $p(W|X, Y)$, we use MC dropout \cite{gal2016dropout}, performing dropout both during training and test time before each weight layer, and approximate the posterior with the simple distribution $q_{\theta}^W$. Here, $q_{\theta}^W$ is a mixture of two Gaussians, where the mean of one of the Gaussians is fixed at zero. We minimize the Kullback-Leibler (KL) divergence between $q_{\theta}^W$ and the $p(W|X, Y)$: ${\cal L}(\theta, p) = \frac{1}{N}\sum_{i=1}^N \log p(y_i|x_i, \theta, X, Y) + \frac{1-p}{2N} ||\theta||^2$, where $N$ is the number of data points, $p$ is dropout probability, $q_{\theta}^W$ is the dropout distribution, and $\hat{W}_t \sim q_{\theta}^W$. Using MC integration with $T$ sampled dropout masks, we have the approximation: $p(y = c|x, X, Y) \approx \frac{1}{T}\sum_{t=1}^{T} \frac{e^{z^{\hat{W}_{c,t}}(x)}}{\sum_{c=1}^{C}e^{z^{\hat{W}_{c\prime, t}}(x)}}$.

Inspired by \cite{malinin2018predictive}, we use entropy in the probability space as a proxy for classification uncertainty. To get an aggregate uncertainty measure, we marginalize over all parameters and use the entropy of the probability vector $p$: $H(p)=-\sum_{c=1}^C p_c \log p_c$. We then quantify the total ($U_t$) and aleatoric uncertainty ($U_a$) using:

\enlargethispage{1\baselineskip}
\vspace{-.5cm}
\begin{align*}
    U_t \approx H[E_{q(\theta|X, Y)}[p(y|x, \theta]]  \approx H[\frac{1}{T}\sum_{t=1}^{T}p(\hat{y}|x, \hat{W}_t)];
\end{align*}

\vspace{-0.3cm}

\begin{align*}
    U_a \approx E_{q(\theta|X,Y)}[H[p(y|x, \theta)]] \approx \frac{1}{T}\sum_{t=1}^{T} H[p(\hat{y}|x, \hat{W}_t)]
\end{align*}


The epistemic uncertainty $U_e$ is then defined $U_t-U_a$. Note that $U_e$ will represent mutual information between true values and model parameters and thus has a different scale compared to $U_a$ and $U_t$ that each represent entropy of a probability distribution. For simplicity, we refer to this model as \textit{UncNet} in the rest of the paper. The code is available at \url{https://github.com/asmadotgh/unc-net}.

\section{Results \& Discussion}

We show that modeling and disambiguating different sources of uncertainty provides a means to identify data that are more difficult to classify, and seek to provide interpretable reasons for why.
Similar to \cite{raghu2018direct}, to represent task subjectivity, we compute the probability that two draws from the empirical histogram of human annotations disagree: $d_i = 1 - \sum_{c=1}^{C}p_{i,c}^2$, where $C$ is the number of classes and $p_{i,c}$ is the probability of image $i$ being rated as class $c$.

\enlargethispage{1\baselineskip}
\subsection{A Proxy for Inter-Rater Disagreement}
Classification of perceived emotions is inherently a subjective task, with disagreement across human annotators. We hypothesize that aleatoric uncertainty is associated with inter-annotator disagreement. We used the Pearson correlation coefficient to assess the relationship between aleatoric uncertainty ($U_a$) and disagreement probability ($d_i$), resulting in a significant correlation: $r = 0.301 , p \ll .001$. This finding suggests aleatoric uncertainty as a tool for quantifying degree of label subjectivity associated with an image.

Note that we observed no significant correlation between epistemic uncertainty and the annotators' disagreement probability: $r=-0.027$, $p=0.105$. This is aligned with our hypothesis that epistemic uncertainty captures the uncertainty introduced by model parameters and is not able to capture the nuance in subjective annotations.
\enlargethispage{1\baselineskip}

\subsection{Task Subjectivity, Difficulty \& Bias in Training}

Figure \ref{fig:extreme-samples} shows samples with the highest and lowest uncertainties. On the left, extreme cases in terms of aleatoric uncertainty ($U_a$) are listed. We observe that samples with low $U_a$ are stereotypical expressions of emotion where annotators (almost) unanimously agree on the assigned label. The fact that ``happiness'' class is the second most common class in the dataset (after ``neutral''), and has a stereotypical morphology in terms of the position of the eye corners, mouth, and teeth exposure may have contributed to the dominance of ``happiness" class in low $U_a$ samples. On the other hand, we observe that samples with highest $U_a$ either represent subjectivity involved in label assignment and lack of annotators' consensus; or low quality of an image. For example, the face occlusion or being a drawing as opposed to a photograph.

Figure \ref{fig:extreme-samples}, on the right, shows extreme cases in terms of epistemic uncertainty ($U_e$). Low $U_e$ samples show similar patterns: samples with stereotypical expression of emotion that are common in the training set. On the other hand, we see different patterns in samples with high $U_e$. We observe that the model has low confidence in the predictions for dark-skinned subgroups. Our interpretation is that the CASIA-WebFace dataset that was used for pre-training the model is highly skewed. It contains faces of celebrities that IMDB lists as active between 1940 and 2014. Most of these celebrities are white. That may explain why the model has high $U_e$ in making a prediction for non-white input images. We also see a sample that exemplifies a non-frontalized photo, which the human annotators were able to unanimously assign a ``neutral'' label despite its atypical viewpoint in the dataset. Since the pre-training process included a frontalization pre-processing step, we believe the current model is not capable of finding meaningful representations for non-frontalized photos and that is why this sample has high $U_e$. Factors such as different illumination may also result in higher $U_e$.




\subsection{A Proxy for Degree of Calibration}

\begin{figure}
\centering  
\includegraphics[width=0.45\linewidth]{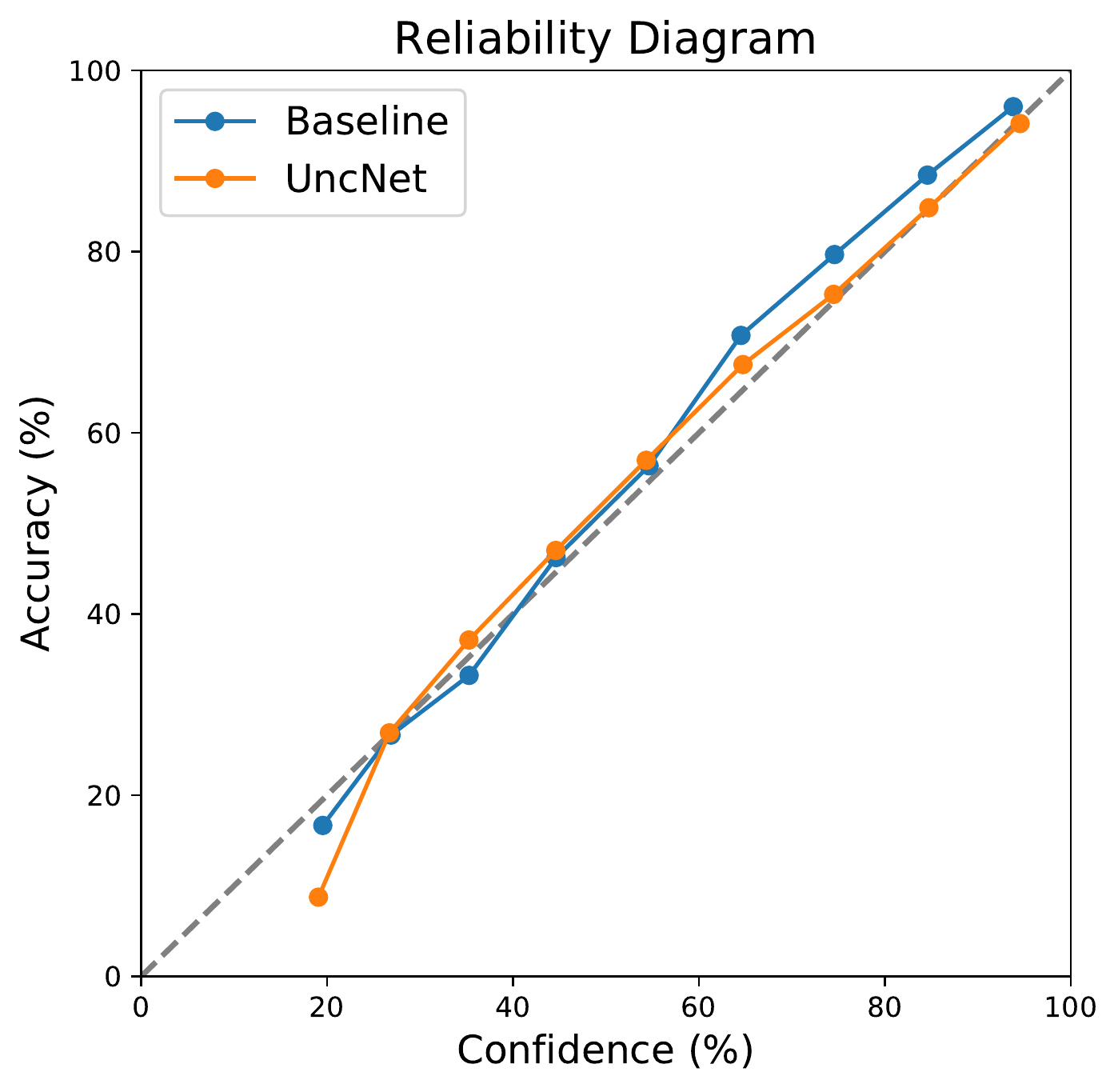}
\hfill
\includegraphics[width=0.45\linewidth]{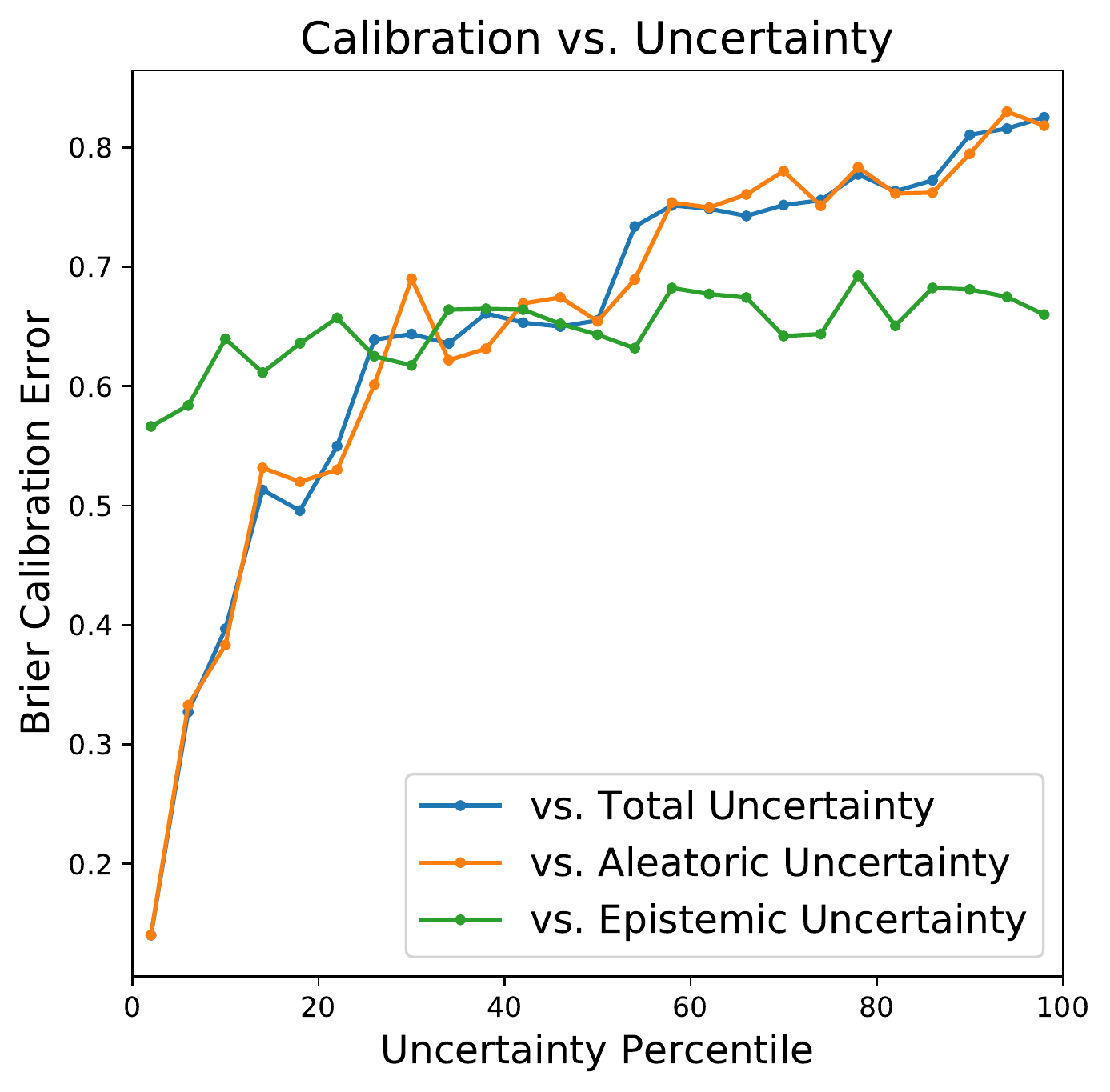}
\caption{\textbf{Left}: Reliability diagram for \textit{Baseline} and \textit{UncNet}. Soft-labels result in well-calibrated predictions. \textbf{Right}: BCE as a function of uncertainty percentile. Uncertainty scores provide details about which sub-group of predictions can be trusted with the confidence scores. Samples with high aleatoric uncertainty are poorly calibrated as measured by BCE. Association between epistemic uncertainty and BCE is in the same direction, but less strong.}
\label{fig:calibration}
\vspace{-.5cm}
\end{figure}



The Brier Calibration Error \cite{brier1950verification, epstein1969verification, lichtenstein1980training} is a commonly-used metric for quantifying calibration: $BCE = \frac{1}{N}\sum_{n=1}^N\sum_{c=1}^C(y_{n, c}-\hat{y}_{n, c})^2$.
$N$ is the number of samples, $C$ is the number of classes, $y$ is a one-hot representation of true labels and $\hat{y}$ is the predicted confidence scores. Additionally, variations of reliability diagram have been used to show the discrepancy between confidence and accuracy \cite{guo2017calibration, niculescu2005predicting,  degroot1983comparison}. Since we have multiple annotations per data point, each pair of {\textless}annotation, sample{\textgreater} is treated separately.

Figure \ref{fig:calibration} (left) shows the reliability diagram for both \textit{Baseline} and \textit{UncNet}. As plotted, both models are close to the 45$^{\circ}$ line. This is aligned with previous research findings showing evidence of well-calibrated predictions when trained with soft-labels \cite{muller2019does, seo2019learning}. While the near-perfect calibration in \textit{Baseline} does not leave space for further improvement, additional uncertainty estimates provide useful information about subgroups of images that may be more or less calibrated. On the right, the relationship between uncertainty and calibration is visualized. We sort samples based on predictive uncertainty estimates and plot BCE as a function of $U_a$, $U_e$, and $U_t$ percentile. There was a significant Pearson correlation between each of these pairs:
$r_{U_a, BCE} = .880, p \ll .00001$; 
$r_{U_e, BCE} = .710, p \ll .00001$;
$r_{U_t, BCE} = .884, p = .00007$.
This suggests that lower uncertainty is associated with a better calibration. Particularly, aleatoric uncertainty plays a more significant role in identifying when a model's predicted confidence score matches the true correctness likelihood. See \S\ref{sec:appendix-calibration} for details regarding other suggested calibration scores \cite{guo2017calibration, nixon2019measuring}.
\enlargethispage{1\baselineskip}

\subsection{Performance}

We also hypothesized performance gains using \textit{UncNet}. Due to task subjectivity and annotation spread (\S\ref{sec:appendix-disagreement}), we believe measures that rely on a binary true/false assumption for evaluation do not fully represent the nuance of our problem setting. Therefore, we use Jensen-Shannon divergence to quantify the distance between predicted and true class probabilities: 
$JSD(p,\hat{p}) =  \frac{KL(p||m) + KL(\hat{p}||m))}{2}$. Here, $m$ is the point-wise mean of $p$ and $\hat{p}$ and $KL$ is the Kullback-Leibler divergence. Lower $JSD(p,\hat{p})$ represents better performance.
A paired-samples t-test was conducted to compare the $JSD$s in \textit{Baseline} and \textit{UncNet}. There was a significant difference in $JSD$ for \textit{Baseline} ($M=0.473, SD=0.131$) and \textit{UncNet} ($M=0.461, SD=0.140$); $t(3578)=9.335, p \ll .001$, confirming our hypothesis.

We take a more granular look and hypothesize that samples with higher uncertainty have higher $JSD(p,\hat{p})$. To test this, a Pearson correlation coefficient was computed to assess the relationship between $U_a$, $U_e$, $U_t$ and $JSD$ in \textit{UncNet}. Each pair showed a significant correlation ($p \ll .00001$):
$r_{U_a, JSD} = .583$; 
$r_{U_e, JSD} = .100$;
$r_{U_t, JSD} = .591$.
This finding further confirms our hypothesis: lower uncertainty is associated with a better match between prediction and groundtruth. Similar to findings of \cite{kendall2017uncertainties}, we see aleatoric uncertainty plays a more significant role in such identification. 

Though accuracy may not fully represent this nuanced problem setting, we also checked how \textit{UncNet} compared to the \textit{Baseline} as measured by accuracy. We observed that \textit{UncNet} has the potential to improve performance modestly, but that if the model had the option to reject classifying samples it is not confident in up to 25\%, it improves significantly in performance, by as much as 8\%. See \S\ref{sec:appendix-performance} for details.


\section{Conclusion}
We focused on the often subjective task of perceived emotion classification and demonstrated how a classical network architecture can be altered to predict measures of epistemic and aleatoric uncertainties and how these measures can help interpretation of model's confidence scores.
We presented evidence for aleatoric uncertainty being a proxy for inter-annotator disagreement and showcased how the measured aleatoric uncertainty can identify low quality inputs or more subjective samples. Additionally, we presented explorations of how epistemic uncertainty can represent bias in training data and suggest directions to alleviate that. Our results suggest that the predicted total uncertainty can act as a surrogate for degree of calibration, even on tasks without human-expert consensus. Finally, we showed there are other benefits such as potential performance improvements.

\enlargethispage{1\baselineskip}

\section*{Acknowledgments}

We would like to thank Ardavan Saeedi and Suvrit Sra for insightful discussions, Jeremy Nixon for sharing calibration metrics code, MIT Stephen A.~Schwarzman College of Computing, Machine Learning Across Disciplines Challenge and MIT Media Lab Consortium for supporting this research.

{
\bibliographystyle{ieee}
\bibliography{references}
}

\appendix

\section{Supplementary Material}

\begin{figure*}[!ht]
  \hspace{-0.5cm}
  \includegraphics[width=1.0\textwidth]{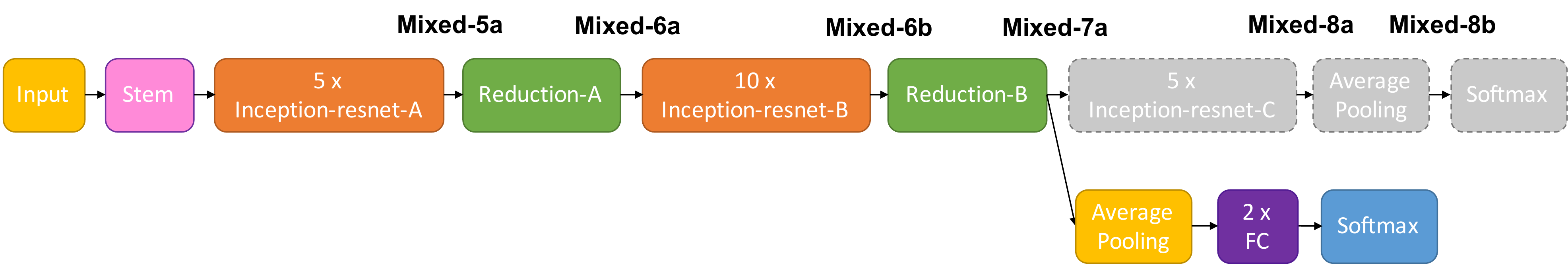}
  \caption{Model architecture: An Inception-ResNet-v1 followed by an average pooling layer and a fully-connected network with two hidden layers (FC). Pre-training on CASIA-WebFace dataset has been conducted on the full Inception-ResNet-V1. We froze the weights of the network and used up to the \texttt{Mixed-7a} layer to extract features from raw images. The remaining unused layers of Inception-ResNet-v1 are in grey. We then stack two FCs on the \texttt{Mixed-7a} layer after average pooling. Dropout is only applied to the FC layers.}
  \label{fig:model}
\end{figure*}

\begin{table*}[!ht]
\caption{Validation accuracy and loss of predicting facial expression emotions on FER+ dataset, using the features extracted from different layers of FaceNet, pre-trained on two different datasets: CASIA-WebFace and VGGFace2.}
\makebox[\textwidth][c]{
\resizebox{\textwidth}{!}{
\centering
\begin{tabular}{@{}llllllll@{}}
\cmidrule(l){2-8}
 &  & \texttt{Mixed-5a} & \texttt{Mixed-6a} & \texttt{Mixed-6b} & \texttt{Mixed-7a} & \texttt{Mixed-8a} & \texttt{Mixed-8b} \\ \cmidrule(l){2-8} 
\multirow{2}{*}{Accuracy (\%)} & CASIA-WebFace & 49.85 & 54.27 & 52.81 & \textbf{55.75} & 52.45 & 52.50 \\
 & VGGFace2 & 50.60 & 54.80 & 55.11 & \textbf{55.50} & 50.69 & 50.46 \\ \cmidrule(l){2-8} 
\multirow{2}{*}{Loss} & CASIA-WebFace & 1.589 & 1.516 & 1.555 & \textbf{1.514} & 1.580 & 1.554 \\
 & VGGFace2 & 1.607 & 1.527 & 1.519 & \textbf{1.513} & 1.589 & 1.598
\end{tabular}
}}
\label{tab:detailedresults}
\end{table*}

\subsection{Model Architecture and Pre-Training Details}
\label{sec:appendix-model}
Figure \ref{fig:model} shows the detailed model architecture. Note that the modules on the top represent an Inception-ResNet-v1 architecture. We have used up to layer \texttt{Mixed-7a} for feature extraction from raw images and added a fully connected (FC) network with two hidden layers of size 128$\times$128. This represents the \textit {Baseline} architecture. The main difference in \textit{UncNet} is adding a dropout mask before each layer of FC, not only during training, but also at inference time.

For face similarity pre-training, we treated the intermediate layer that was used to export features from a raw image as a hyper-parameter that was tuned according to validation loss. Our experiments included \texttt{Mixed-5a}, \texttt{Mixed-6a}, \texttt{Mixed-6b}, \texttt{Mixed-7a}, \texttt{Mixed-8a}, and \texttt{Mixed-8b} layers. We observed that the \texttt{Mixed-7a} layer best encoded and retained emotional information from the input face crop. Table \ref{tab:detailedresults} summarizes our exploration results.

\subsection{Annotation Disagreement Details}
\label{sec:appendix-disagreement}

\begin{figure}[!t]
  \includegraphics[width=0.95\columnwidth]{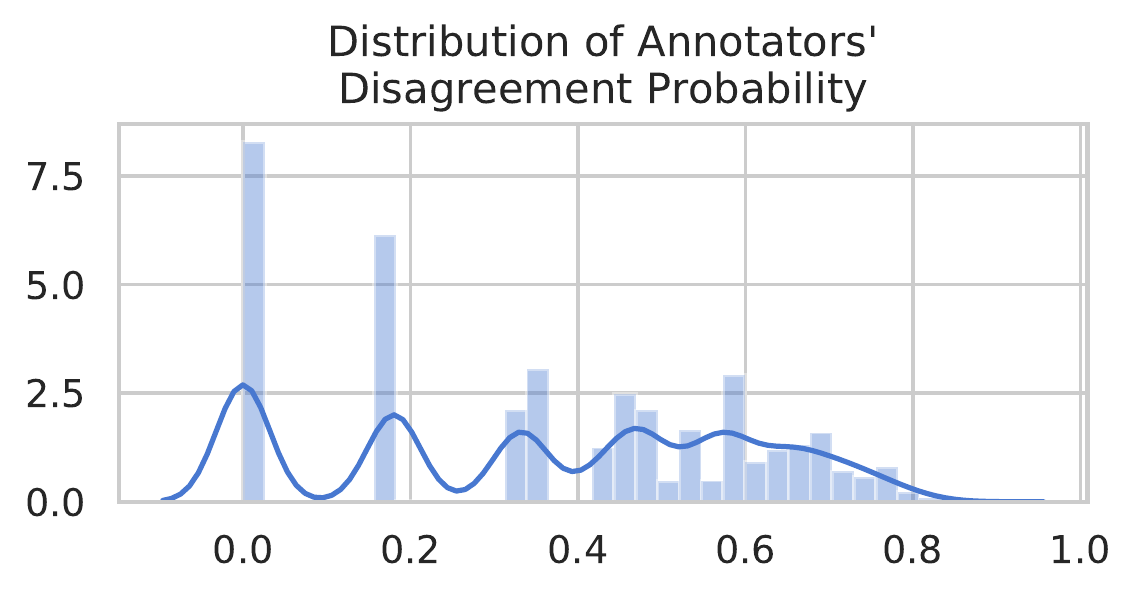}
  \caption{Distribution of annotators' disagreement probability ($d_i$) on FER+ training samples. The histogram heights are scaled to represent density rather than absolute count, so that the area under the fitted curve is one.}
  \label{fig:annotator-disagreement}
\end{figure}

Figure \ref{fig:annotator-disagreement} shows the distribution of disagreement probability ($d_i$) for all images in the training set. Histogram heights show a density rather than the absolute count, so that the area under the fitted curve is one.

\subsection{Detailed Calibration Results}
\label{sec:appendix-calibration}

In this section, we report a range of calibration scores for \textit{Baseline} and \textit{UncNet}. Further, we show how these scores are related to the predictive uncertainty estimates of \textit{UncNet}. 

Scholars have introduced a range of calibration scores. Maximum Calibration Error (MCE) and Expected Calibration Error (ECE) approximate calibration error by quantization of uncertainty bins and have been adopted in many recent publications \cite{guo2017calibration}:
\begin{equation*}
    MCE = \max_{b=1}^B |acc(b) - conf(b)|
\end{equation*}
\begin{equation*}
    ECE = \sum_{b=1}^B \frac{n_b}{N} |acc(b) - conf(b)|
\end{equation*}

Here $n_b$ is the number of predictions in bin $b$, $N$ is the number of samples, $acc(b)$ is the accuracy of prediction in bin $b$, and $conf(b)$ is the average prediction confidence score in bin $b$. Recently, new metrics have been proposed to overcome the limited assumption of mutually exclusiveness of classes and improve robustness to label noise \cite{nixon2019measuring}. Static Calibration Error (SCE) is a metric where prediction for all classes is taken into account as opposed to only the argmax of softmax outputs. Adaptive Calibration Error (ACE) is an extension of SCE where instead of equidistant bins, confidence scores are sorted and their percentiles represent ``ranges", parallel to ``bins" in SCE. Thresholded Adaptive Calibration Error (TACE) is an extension to ACE where values with at least $\epsilon$ confidence are taken into account. SCE, ACE, and TACE can be formally defined as the following:

\begin{equation*}
    SCE = \frac{1}{C}\sum_{c=1}^{C}\sum_{b=1}^B\frac{n_{bc}}{N}|acc(b,c) -conf(b, c)|
\end{equation*}
\begin{equation*}
    ACE = \frac{1}{CR}\sum_{c=1}^C\sum_{r=1}^R|acc(r,c)-conf(r,c)|
\end{equation*}

\begin{align*}
\begin{split}
TACE & = \frac{1}{CR}\sum_{c=1}^C\sum_{r \in R_{\prime}} |acc(r,c)-conf(r,c)| \\
& where \quad \forall r \in R_{\prime}: conf(r,c)>\epsilon
\end{split}
\end{align*}

\begin{table}[!h]
\caption{Summary of additional calibration error metrics for \textit{Baseline} vs. \textit{UncNet}. Near-perfect calibration with soft-labels and dependency of these metrics on quantization may be potential reasons for having inconclusive results.}
\makebox[\columnwidth][c]{
\resizebox{\columnwidth}{!}{
\begin{tabular}{@{}lllllll@{}}
\toprule
\multicolumn{1}{c}{\begin{tabular}[c]{@{}c@{}}Calib.\\ Error\end{tabular}} & \multicolumn{1}{c}{\begin{tabular}[c]{@{}c@{}}ECE\\ (\%)\end{tabular}} & \multicolumn{1}{c}{\begin{tabular}[c]{@{}c@{}}MCE\\ (\%)\end{tabular}} & \multicolumn{1}{c}{\begin{tabular}[c]{@{}c@{}}SCE\\ (\%)\end{tabular}} & \multicolumn{1}{c}{\begin{tabular}[c]{@{}c@{}}ACE\\ (\%)\end{tabular}} & \multicolumn{1}{c}{\begin{tabular}[c]{@{}c@{}}TACE\\ (\%)\end{tabular}} & \multicolumn{1}{c}{BCE} \\ \midrule
\textit{Baseline} & 2.330 & \textbf{6.231} & 0.479 & \textbf{0.390} & \textbf{0.393} & 0.661 \\
\textit{UncNet} & \textbf{1.876} & 10.326 & \textbf{0.417} & 0.526 & 0.506 & \textbf{0.649} \\ \bottomrule
\end{tabular}
}
}
\label{tab:detailed-calibration}
\end{table}

Table \ref{tab:detailed-calibration} summarizes these metrics using B/R=10. We did not observe any conclusive results comparing \textit{Baseline} and \textit{UncNet} conditions or using uncertainty quantiles. Our interpretation is that the close-to-perfect calibration with soft-labels, as well as identified problems with the dependence of these metrics on quantization may have resulted in a null result. Further study in this area is required to better understand what these metrics can and cannot capture.

\subsection{Detailed Performance Metrics}
\label{sec:appendix-performance}
In this section, we report accuracy for a random run on the test set. Accuracy is defined as the percentage of samples where predicted maximum probability class maps to the annotated maximum probability class. Table \ref{tab:performance} summarizes our findings.
For future, we will add further performance metrics such as average precision or per-class accuracy and provide confidence bounds using bootstrapping.
\begin{table}[ht]
\caption{Summary of performance metrics for \textit{Baseline} vs.~\textit{UncNet} and how it is influenced if given the possibility of rejecting classification of certain samples. $U_e$: Epistemic uncertainty, $U_a$: Aleatoric uncertainty, $U_t$: Total uncertainty.}
\centering
\begin{tabular}{@{}lll@{}}
\toprule
\textbf{Model} & \textbf{Evaluation Dataset} & \textbf{Accuracy (\%)} \\ \midrule
\textit{Baseline} & FER+ Test & 54.848 \\
\textit{UncNet} & FER+ Test & 56.943 \\
\textit{UncNet} - low $U_e$ & 75\% of FER+ Test & 57.452 \\
\textit{UncNet} - low $U_a$ & 75\% of FER+ Test & 62.481 \\
\textit{UncNet} - low $U_t$ & 75\% of FER+ Test & 62.332 \\ \bottomrule
\end{tabular}
\label{tab:performance}
\end{table}

\end{document}